# Securing Pathways with Orthogonal Robots

Hamid Hoorfar
Department of Epidemiology and Public Health
University of Maryland
Baltimore, Maryland, USA
hhoorfar@som.umaryland.edu

Faraneh Fathi
Department of Biomedical Engineering
University of Kentucky
Lexington, Kentucky, USA
faraneh.fathi@uky.edu

Sara Moshtaghi Largani
Department of Computer Science
University of Cincinnati
Cincinnati, Ohio, USA
moshtasa@mail.uc.edu

Alireza Bagheri
Department of Computer Engineering
Amirkabir University of Technology
Tehran, Iran
ar.bagheri@aut.ac.ir

*Abstract*— The protection of pathways holds immense significance across various domains, including urban planning, transportation, surveillance, and security. This article introduces a groundbreaking approach to safeguarding pathways by employing orthogonal robots. The study specifically addresses the challenge of efficiently guarding orthogonal areas with the minimum number of orthogonal robots. The primary focus is on orthogonal pathways, characterized by a path-like dual graph of vertical decomposition. It is demonstrated that determining the minimum number of orthogonal robots for pathways can be achieved in linear time. However, it is essential to note that the general problem of finding the minimum number of robots for simple polygons with general visibility, even in the orthogonal case, is known to be NP-hard. Emphasis is placed on the flexibility of placing robots anywhere within the polygon, whether on the boundary or in the interior.

I. INTRODUCTION

Pathway protection and preservation play a pivotal role in numerous domains, such as urban planning, transportation, surveillance, and security. Here, we present a revolutionary method for safeguarding pathways, utilizing orthogonal robots. Our research focuses on effectively securing uncomplicated orthogonal regions using the fewest possible orthogonal robots. The key aspect of our approach involves delving into the concept of visibility within the realm of orthogonal polygons. Visibility refers to the ability to observe or detect a point from a specific location. In this study, the visibility of an interior point p within an orthogonal polygon P is determined based on whether the minimum area axis-aligned rectangle containing both p and a guard point q lies entirely within P [1-3]. A guard set (G) is defined as a collection of point guards within the orthogonal polygon P. The objective of the guard set is to ensure that the union of their visibility areas covers the entire polygon, thereby making every point within P visible from at least one guard in G. The primary focus of this study is on orthogonal pathways, which are characterized by a path-like dual graph of vertical decomposition. By leveraging the path-like structure of these pathways, the study aims to determine the minimum number of orthogonal robots required to effectively guard them. A key advantage of this approach is its ability to accomplish this task in linear time, offering a practical and efficient solution. It is worth mentioning previous work by Worman and Keil [18], who developed an algorithm with a time complexity of polynomial time to address the problem of guarding orthogonal path polygons. However, it is important to note that the general problem of finding the minimum number of guards for simple polygons with general visibility, even in the orthogonal case, is known to be NP-hard. One notable aspect emphasized in this article is the flexibility of guard placement within the polygon. Guards can be positioned anywhere within the orthogonal polygon, whether on the boundary or in the interior.

*A. Related works*

Every year, a multitude of papers are published in various scientific fields related to algorithm, encompassing a wide range of topics such as robotics [4, 5], big data [6], social network [7, 8], sampling [9], graph theory [10-12], artificial intelligent, deep learning, smart city facility [13] and communication system [14] as well as security and guarding. These papers specifically focus on advancing the development of new algorithms tailored to enhance security and guarding. The art gallery problem targets the identification of a set G of point guards within a polygon P, such that every point in P is visible from at least one guard in G. In this problem, a guard g and a point p are considered visible if the line segment $gp$ is contained entirely within the polygon [15, 16]. It has been proven that finding the optimal number of guards (minimum guard set) required to cover an arbitrary simple polygon is NP-hard [15]. The complexity of the art gallery problem persists even for orthogonal polygons and monotone polygons [17]. An orthogonal polygon is characterized by edges that are either horizontal or vertical, with the number of vertical edges equaling the number of horizontal edges. Worman and Keil [18] investigated the decomposition of orthogonal polygons into the optimum number of r-star (star-shaped) sub-polygons, which is equivalent to solving the orthogonal art gallery problem. They presented a polynomial-time algorithm for this problem under orthogonal visibility (so-called r-visibility), demonstrating its polynomial solvability. However, their algorithm has a time complexity of $O(n^{17} poly\ log\ n)$, which limits its efficiency. A cover of a polygon P is achieved when a set S of sub-polygons is selected in such a way that their union exactly forms the polygon P. An r-star polygon refers to an orthogonal star-shaped polygon. The task of determining a minimum cover for a simple orthogonal polygon using r-stars is directly equivalent to finding the minimum set of orthogonal guards necessary to ensure visibility of every point within the polygon. This equivalence highlights the direct relationship between polygon coverage and the protection of pathways, emphasizing the practical relevance of the cover problem in the context of pathway safeguarding. In their work, Gewali et al. [19] introduced a linear-time algorithm specifically designed to cover monotone orthogonal polygons with the minimum number of r-star polygons. This algorithm offers an efficient solution to the problem of achieving coverage in monotone orthogonal polygons. Similarly, Palios and Tzimas [20] focused their research on a specific class of orthogonal polygons that do not contain any holes. These polygons have reflex edges (also known as dents) along at most three different orientations. The authors developed an algorithm with a time complexity of $O(n + k \log k)$, where n represents the size of the polygon and k denotes the size of a minimum r-star cover. This algorithm provides an effective approach for determining the minimum set of r-star polygons needed to achieve complete coverage within this class of orthogonal polygons. In addition, Beidl and Mehrabi [21] provided evidence that determining a minimum cover on orthogonal polygons with holes is an NP-hard problem. This finding highlights the computational complexity associated with finding optimal coverings in such polygonal configurations.

Despite the notable progress made in this field, the challenge of determining the minimum number of robots required for various environment configurations remains formidable, even when focusing solely on the fixed positions of each robot rather than their mobility. Continuing efforts in research are dedicated to devising more efficient algorithms and delving into the practical applications of these solutions. The ultimate goal of this ongoing research is to tackle the intricacies involved and enhance our capacity to effectively safeguard pathways in diverse real-world scenarios. It's important to note that in parallel with this research, there are others exploring the positions of robots in the environment. Some researchers may be dedicated to finding optimal paths for the robots to navigate and cover the entire environment efficiently [22, 23].

*B. Motivation*

Ensuring the safety and protection of areas is a critical concern in various domains, including urban planning, transportation, surveillance, and security. Pathways within these areas serve as vital conduits for human movement, transportation networks, and the flow of goods and services. However, these pathways can also be vulnerable to potential threats, such as unauthorized access, intrusions, or security breaches. Therefore, developing effective methods to secure and safeguard these areas is of paramount importance. Traditional approaches to area security often rely on human personnel or fixed surveillance systems, which may have limitations in terms of coverage, response time, and adaptability. This motivates the exploration of novel technological solutions that can enhance the security of areas and overcome these limitations.

In recent years, advancements in robotics and autonomous systems have paved the way for innovative approaches to area security. Robotic guardians, equipped with advanced sensors, autonomous navigation capabilities, and intelligent decision-making algorithms, have the potential to revolutionize the field of area protection. By leveraging the capabilities of these orthogonal robotic guardians, it becomes possible to enhance the monitoring and safeguarding of areas in a more efficient, flexible, and adaptive manner.

The motivation behind this paper stems from the need to address the challenges associated with securing areas using orthogonal robotic guardians. While the problem of determining the minimum number of guards for simple areas with general visibility is known to be computationally complex, our research aims to develop a geometric approach and propose a novel strategy for efficient guard placement within orthogonal areas. By emphasizing the flexibility of guard deployment, both on the boundary and in the interior of the area.

TABLE I - LIST OF POLYGONS DESCRIPTION

| Object | Description |
| --- | --- |
| Orthogonal polygon | a polygon whose edges are aligned with the coordinate axes (x-axis and y-axis) and consist of segments that are either horizontal or vertical. |
| pixel | A rectangle part after decomposition |
| tree polygon [21] | a polygon is considered a "tree polygon" if its dual graph is a tree. |
| k-width polygon [20] | a polygon is labeled "k-width" if its dual graph is a k-width tree. |
| Thin polygon [21] | If all the vertices of the pixels lie on the boundary of P after the decomposition, the polygon is referred to as "thin." |
| path polygon | An orthogonal polygon is termed a "path polygon" if the dual graph of its vertical decomposition (not the general decomposition) is a path. |
| tooth edge | An edge with two endpoints of angle $\pi/2$ |
| dent edge | An edge with two endpoints of angle $3\pi/2$ |
| x-monotone polygon | An orthogonal polygon has no dent edges in the directions perpendicular to the y-axis. |
| ortho-convex polygon | An orthogonal polygon with no dent edges |
| star-shaped polygon | A polygon P has an internal point from which the entire P is straight-line visible. |
| r-star polygon | An orthogonal polygon P that can be seen (r-visible) from a internal point, and the entire P is visible from it. |
| kernel | The set of internal points that provide visibility to the entire polygon is referred to as the kernel. |
| bounding box | the bounding box of a polygon P is the minimum area rectangle aligned with the axes, containing all the points of P. |
| Histogram polygon | An x-monotone (y-monotone) orthogonal polygon shares a horizontal (vertical) edge with its bounding box and the shared edge is called the base. |
| pyramid polygon | an ortho-convex polygon shares an edge with its bounding box. |
| fan polygon | an ortho-convex polygon that shares two adjacent edges with its bounding box. |

In the next section, we undertake mathematical modeling of the guarding concept. To demonstrate the accuracy and correctness of our algorithms, we introduce notations that prove invaluable. The pathways are represented as path polygons, while the robots are depicted as observable points. Therefore, the problem at hand is transformed into the task of identifying the minimum number of observable points from which the entire polygon can be visible. Through the utilization of these specific notations, we establish a vital framework for substantiating the precision and validity of our algorithms. In this paper, visibility means r-visibility (orthogonal visibility) and guarding is under r-visibility and also, monotonicity means x-monotonicity unless explicitly mentioned.

## II. PRELIMINARIES

An orthogonal polygon, denoted as P with n vertices, can be decomposed by extending the edges of P that are incident to their reflex vertices until they intersect the boundary. For explanations of specific area names, please refer to TABLE I. As a result of this process, a maximum of $(\frac{n}{2} - 1)^2$ pixels are created. By assigning a node to each pixel and connecting nodes whose corresponding pixels are adjacent, we create a graph known as the dual graph. The type of dual graph determines the classification of the orthogonal polygon. A vertical decomposition is obtained by extending only the vertical edges. This type of partition, called a vertical partition, results in the creation of at most $n/2 - 1$ rectangles. The edge direction is defined as the direction of its normal vector from the interior to the exterior of the polygon. The class-i of an orthogonal polygon contains polygons that have dent edges in only i different directions [20]. Consider vertical decomposing path polygon, the resulting dual graph G of this vertical decomposition has two nodes of degree one. The rectangles corresponding to these two nodes are called the "first rectangle" or "last rectangle". Let R = $R_1, R_2, ..., R_m$ be the series of the rectangles, where m = n-2, ordered from the first to the last rectangle according to the order of their corresponding nodes in the graph G. For illustration, refer to Fig. 1, where $R_1$ and $R_{26}$ represent the first and last rectangles in this example, respectively.

TABLE II should be referred to for the explanations of the mathematical notations. Without loss of generality, we assume that for any two different vertical edges e and e', $x(e) \neq x(e')$. A member of R is labeled as a local minimum if its height is less than its two adjacent rectangles. In a polygon, two objects o and o' are defined as weak visible if every point of o is visible to at least one point of o'. The problem of decomposing an orthogonal polygon into the minimum number of r-star sub-polygons is equivalent to the problem of guarding an orthogonal polygon with the minimum number of r-guards, as discussed in the previous section.

**Lemma 1:** An ortho-convex polygon P is an r-star polygon if the leftmost and rightmost vertical edges of P are weakly visible to each other, and the upper and lower horizontal edges of P are weakly visible to each other.

**Proof.** Since the leftmost and rightmost vertical edges of P are weakly visible, there exists a horizontal line segment $\delta_1$ connecting the leftmost and rightmost vertical edges of P, which lies entirely within P. Similarly, since the upper and lower horizontal edges of P are weakly visible, there exists a vertical line segment $\delta_2$ connecting the upper and lower horizontal edges of P, which lies entirely within P. As $\delta_1$ connects the leftmost and rightmost vertical edges of P and $\delta_2$ connects the upper and lower horizontal edges of P, they must intersect at a point M that is contained within P. $\delta_2$ and $\delta_2$ divide P into four parts; $P_1, P_2$,

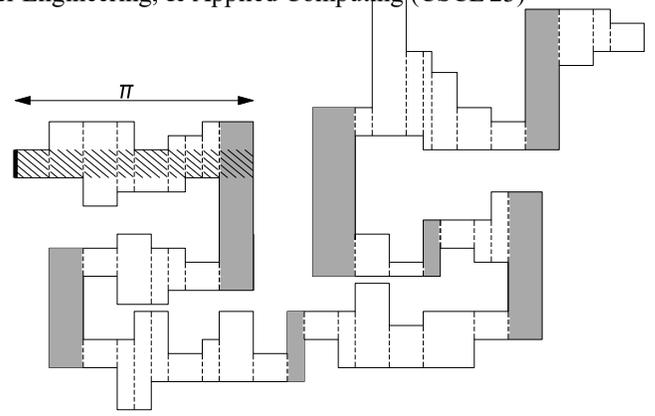

Fig. 1 - Decomposition of a path polygon into balanced parts

TABLE II - LIST OF MATHEMATICAL NOTATIONS

| Notation | Parameter Type | Description |
| --- | --- | --- |
| R | - | the series of the rectangles, after vertical decomposition, ordered |
| Int(P) | polygon | Interior of polygon P |
| Ext(P) | polygon | Exterior of polygon P |
| Bound(P) | polygon | Boundary of polygon P |
| x(p) | Point/vertical segment | x-coordinate of p |
| y(p) | Point/horizontal segment | y-coordinate of p |
| e(s) | Line segment | the horizontal edge of P that contains the segment s |
| $u_i$ | - | the upper horizontal edges of rectangle $R_i$ |
| $l_i$ | - | the lower horizontal edges of rectangle $R_i$ |
| $s_i$ | - | the shared vertical segment between $R_i$ and $R_{i+1}$ |
| $E_U$ | - | the series of horizontal edges of the upper chain of P, ordered |
| $E_L$ | - | the series of horizontal edges of the lower chain of P, ordered |
| U | - | the series of horizontal edges of the upper edges of R, ordered |
| L | - | the series of horizontal edges of the upper edges of R, ordered |
| left(e) | Horizontal edge | The left endpoint of e |
| right(e) | Horizontal edge | The right endpoint of e |
| top(e) | Vertical edge | The top endpoint of e |
| down(e) | Vertical edge | The bottom endpoint of e |
| $os_e$ | horizontal edge | The set of points p in the polygon such that there exists a point q $\in$ e, and the line segment pq is perpendicular to e and entirely contained within the polygon, is referred to as the orthogonal shadow of e |

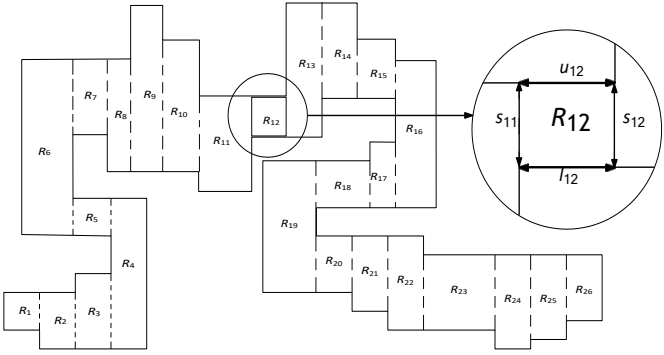

Fig. 2 - vertical decomposition of pathways (path polygons)

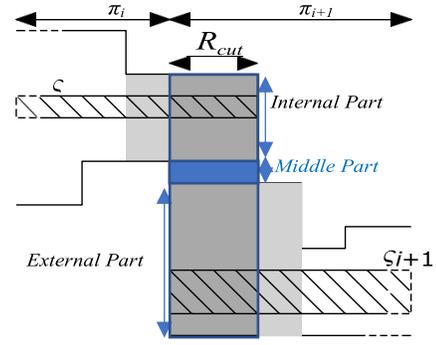

Fig. 3 - cut rectangle ( R-cut)

$P_3$, and $P_4$. All of these parts are fan polygons and share M as kernel. ∎

In the next section, we present a linear-time exact algorithm for finding the minimum guarding of path polygons (pathways).

## III. ALGORITHM FOR GUARDING PATHWAYS

A path polygon displays a notable property that enables its division into multiple sub-polygons, where the shortest route for a single robot within these sub-polygons follows an orthogonal straight-line segment. For each sub-polygon, an optimal set of guards exists where all guards are positioned along the shortest watchman route of that particular sub-polygon. By leveraging this insight, we can identify an optimal guard set that resides on a set of line segments, thereby reducing the execution time of the algorithm. Additionally, we can establish that the visibility areas of guards located within one sub-polygon do not intersect with the visibility areas of guards positioned in other sub-polygons. Consequently, the minimum number of guards required to protect the path polygon will be equal to the sum of the minimum numbers of guards needed for the individual sub-polygons. These sub-polygons are referred to as balanced orthogonal polygons, as they are both monotone and walkable along a straight-line path. In this context, walkable denotes that a mobile robot can effectively cover the entire polygon by traversing back and forth along a straight route. Essentially, these polygons encompass an area known as a corridor, within which a straight shortest robot path segment is included. For instance, consider a histogram polygon, where its base edge serves as a robot route that constitutes part of its corridor. In the upcoming subsection, we will employ a ray-shooting (beam throwing) technique to determine this corridor. It is important to note that a path polygon is not inherently walkable in a straight manner (or balanced). Therefore, we will decompose the path polygon into the minimum number of balanced parts and subsequently assign guards to each part individually. Initially, the path polygon falls under class-4 of the aforementioned orthogonal classification, but after this decomposition, all the resulting parts belong to class-2. This reclassification occurs because all the dent edges of the path polygon, which possess horizontal directionality, are eliminated during the partitioning process.

### A. Decomposition of a pathway into Balanced Parts

Let P be a path polygon with n vertices, and let R, U, L be its related series, as defined in the previous section. The first rectangle and the general path polygon share a common vertical edge, denoted as ε. To decompose the path polygon, we propagate a light beam along a rectilinear direction perpendicular to ε and collinear with the X-axis. The light beam, or a portion of it, passes through a subset of R (referred to as $R_\pi$), creating a sub-polygon π of P (see Fig. 1)). The rectangles belonging to sub-polygon π possess a geometric property where the y-coordinates of their upper edges are greater than the y-coordinates of their lower edges. In other words, in sub-polygon π, all the dent edges of the upper chain are higher than those of the lower chain. Therefore, there exists a rectangular corridor ς that connects the leftmost and rightmost vertical edges of π, with no intersection with ext(π). If we denote the leftmost vertical edge of sub-polygon π as v and the rightmost vertical edge as v'.

$$y_1 = min_{u_i \in \pi}(y(u_i)) , y_2 = max_{l_i \in \pi}(y(l_i)) \qquad (1)$$

ς is an axis-aligned rectangle spanned by two points with coordinates (x(v), $y_1$) and (x(v'), $y_2$). Consequently, π is both walkable and balanced, and any horizontal line segment connecting v and v' within ς can serve as its shortest watchman route. After identifying the first balanced sub-polygon (part) π, we remove it from the path polygon P and repeat these operations to find the next balanced parts until P is completely decomposed into balanced and monotone parts. TABLE III should be referred to for the explanations of the mathematical notations.

TABLE III. LIST OF BALANCED PARTS COMPONENTS

| Notation | Parameter Type | Description |
|---|---|---|
| $\pi_1, \pi_2, ..., \pi_k$ | - | the series of all the balanced parts, after decomposition, ordered |
| $R_{\pi_i}$ | balanced part $\pi_i$ | the subseries of R located in $\pi_i$ |
| $\varsigma_i$ | Index i of a balanced part | Corridor of $\pi_i$ |
| R-cut | - | The last rectangle of $R_{\pi_i}$ |
| R-cut-i | Offset i | The i-th previous rectangle of R-cut |
| R-cut+i | Offset i | The i-th next rectangle of R-cut |
| H(A) | Rectangle A | The height of A |

However, it is crucial to clarify an important aspect regarding the last rectangles of each obtained balanced sub-polygon (part). A rectangle, which represents the border area between two parts denoted as R-cut, is referred to as the last rectangle in every balanced part (Fig. 4). It can belong to either $\pi_i$ or $\pi_{i+1}$. When aiming to minimize the number of guards required to guard the path polygon, a distinction arises between two cases: R-cut belonging to $\pi_i$ (case 1) or R-cut belonging to $\pi_{i+1}$ (case 2). In the case where R-cut-1 represents a local minimum, it is optimal to assign R-cut to $\pi_{i+1}$ (case 2). We prove this proposition in lemma 2.

Each cut rectangle can be divided into three distinct parts by extending the horizontal edges of R-cut-1 and R-cut+1, which intersect the boundary at their common vertices with R-cut. These parts are referred to as internal, middle, and external parts. An illustration of this can be seen in Fig. 3. The internal part is adjacent to R-cut-1, the external part is adjacent to R-cut+1, and the middle part is the third part. It is essential to place a guard within the R-cut to ensure its coverage. This is because it is impossible for the interior of the middle part to be guarded by an r-guard that is not situated within R-cut. If the previous rectangle R-cut-1 is a local minimum, we remove R-cut from $R_{\pi_i}$ and assign it to $R_{\pi_{i+1}}$. By this strategy, the number of required guards can be reduced in certain cases.

**Claim 1:** There exists a minimum guard set of every path polygon, where all guards are positioned within its corridors.

**Lemma 2:** Optimally, for guarding the path polygon P, it is preferable to assign R-cut to $\pi_{i+1}$ rather than assigning it to $\pi_i$ if R-cut-1 is a local minimum.

**Proof.** While the interior of the middle part of R-cut is not visible from any points in P – R-cut, it is necessary to place a guard g inside R-cut to guard it. We need to find an optimal position for the guard as denoted by g. There are two possibilities: placing g in R-cut ∩ $\varsigma_i$ or R-cut ∩ $\varsigma_{i+1}$. We need to determine which choice leads to the minimum guarding of the path polygon P. Figure 4 illustrates four different cases.

In cases (a) and (b), R-cut−1 is a local minimum. If we place g in R-cut ∩ $\varsigma_i$, then R-cut−1 is guarded, but R-cut−2 is not completely guarded. Thus, there must be a guard g' in the guard set that covers R-cut−2. Additionally, while H(R-cut−2)> H(R-cut−1), g' can also fully guard R-cut−1. Therefore, if R-cut−1 is a local minimum, placing g in R-cut ∩ $\varsigma_i$ is not advantageous. Consequently, it is better to place g in R-cut ∩ $\varsigma_{i+1}$. This occurs when we assign R-cut to $\pi_{i+1}$ instead of assigning it to $\pi_i$ (regardless of whether R-cut+1 is a local minimum or not).

In case (c), R-cut+1 is a local minimum. If we place g in the area R-cut ∩ $\varsigma_{i+1}$, both R-cut and R-cut+1 are guarded, but R-cut+2 is not fully guarded. Therefore, there must be a guard g' in the guard set that covers Rcut+2. Moreover, while H(R-cut+2)>H(R-cut+1), g' can also completely guard R-cut+1 (g' is located somewhere in $\varsigma_{i+1}$). Thus, if R-cut+1 is a local minimum, placing g in R-cut ∩ $\varsigma_{i+1}$ is not advantageous. It is better to place g in R-cut ∩ $\varsigma_i$. This occurs when we assign R-cut to $\pi_i$ instead of assigning it to $\pi_{i+1}$, while R-cut−1 is not a local minimum.

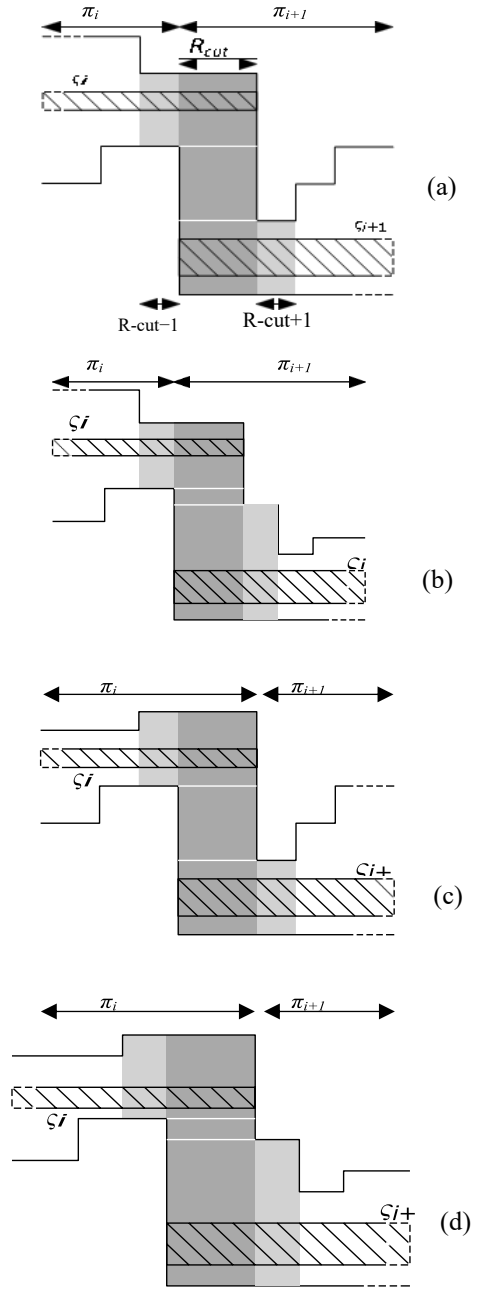

Fig. 4 - Four different cases occur for assigning R-cut

In case (d), both R-cut−1 and R-cut+1 are not local minimum. R-cut−1 is not a local minimum, and its height is greater than that of R-cut−2. Therefore, to guard R-cut−1, it is necessary to place a guard in (R-cut−1 ∪ R-cut) ∩ $\varsigma_i$. Similarly, R-cut+1 is not a local minimum, and its height is greater than that of Rcut+2. Consequently, to guard R-cut+1, a guard must be placed in (R-cut ∪ R-cut+1) ∩ $\varsigma_{i+1}$. However, it is unnecessary to have two guards within R-cut. For the sake of simplicity, we place one guard in R-cut ∩ $\varsigma_i$ and another guard in R-cut+1 ∩ $\varsigma_{i+1}$. This arrangement occurs when we assign R-cut to $\pi_i$ instead of assigning it to $\pi_{i+1}$. The four cases described in Fig. 4 depict adjacent rectangles R-cut−1 and R-cut+1 located on different sides of R-cut. If Rcut−1 and Rcut+1 are on the same side, four additional cases arise that are similar to the previous four cases. ∎

When decomposing the path polygon P into balanced sub-polygons, we follow a step-by-step approach. We begin by finding the first balanced sub-polygon of P. After its identification, we remove it from P and proceed to iterate the algorithm on the remaining part of P, denoted as π. This process continues until P is fully decomposed. During the decomposition, we remove the rectangles belonging to π from set R. The members of R are ordered and labeled starting from 1. After removal, we relabel the remaining members from 1 again, simplifying the algorithm's description. The same procedures are carried out for the series U and L. The number of iterations required is equal to the cardinality of R at the beginning. Hence, the time complexity of decomposing the path polygon P into balanced sub-polygons is linear, directly influenced by the size of P. The pseudocode of the algorithm is presented as Algorithm 1.

Let's assume that P has been decomposed into a set of balanced sub-polygons, $\pi_1, \pi_2, ..., \pi_k$, and their respective corridors are $\varsigma_1, \varsigma_2, ..., \varsigma_k$. It can be observed that if $i \neq j$, there is no point within the interior of $\varsigma_i$ that is orthogonally visible from $\varsigma_j$. This property allows us to optimally cover P by placing guards exclusively on the corridors. In this approach, guarding each sub-polygon can be done independently. Therefore, the minimum number of guards required to secure the entire polygon is equal to the sum of the minimum number of guards needed for each sub-polygon. It is important to note that claim 1 has to be proven. In the subsequent sections, we will present an algorithm that supports the claim, and we will demonstrate its optimality through rigorous proof.

*B. The Algorithm for Guarding the Balanced Sub-polygons*

In the previous subsection, we established that every balanced (walkable) polygon possesses a rectangular area known as a corridor, from which the entire polygon is weakly visible. In this section, we present an algorithm for determining the minimum number of guards and their respective positions within the corridor for a balanced polygon. Consider a polygon P that is both orthogonal and monotone, with n vertices and its related series. Additionally, we identify ε as the leftmost vertical edge of P and ε' as the rightmost vertical edge. Furthermore, $e_{min}$ represents the lowest horizontal edge of the upper chain, while $e_{max}$ corresponds to the highest horizontal edge of the lower chain. The corridor of P, spanned by the points $(x(\varepsilon), y(e_{min}))$ and $(x(\varepsilon'), y(e_{max}))$, is guaranteed to be non-empty.

**Lemma 3:** Every tooth edge t can be covered by a guard placed only in its orthogonal shadow $os_t$.

**Proof.** suppose the tooth edge t is guarded by a guard γ that is not placed in $os_t$. This implies that γ does not lie on the x-coordinate of any point on the edge t, without loss of generality assume $x(Left(t)) > x(\gamma)$. while γ is visible to t, both endpoints, Right(t) and Left(t), are visible to γ. Thus, there exists an axis-aligned rectangle spanned by Right(t) and γ that is contained within the polygon. Since the vertices of the rectangle, Rt = (x(Right(t)), y(Right(t)), A = (x(Right(t)), y(γ)), B = (x(γ), y(Right(t))), and g = (x(γ), y(γ)), all belong to the polygon, the horizontal edge between B and Right(t) is entirely contained within the polygon. This is a contradiction because t is not part of the edge between B and Right(t). If an edge of a polygon is a subset of a longer line segment that belongs to the polygon, it is not considered a real edge. ∎

**Data:** an path polygon with n vertices
**Result:** minimum number of balanced monotone polygons
set $min_u = u_1$ and $max_l = l_1$;
**while** set of rectangles R is not empty **do**
    **if** $u_i > max_l$ or $l_i < min_u$ **then**
        **if** $i - 2 = 1$ or $R_{i-2}$ is not local minimum **then**
            $R = R - \{R_1, R_2, ..., R_{i-2}, R_{i-1}\}$;
            $U = U - \{u_1, u_2, ..., u_{i-2}, u_{i-1}\}$;
            $L = L - \{l_1, l_2, ..., l_{i-2}, l_{i-1}\}$;
        **else**
            $R = R - \{R_1, R_2, ..., R_{i-2}\}$;
            $U = U - \{u_1, u_2, ..., u_{i-2}\}$;
            $L = L - \{l_1, l_2, ..., l_{i-2}\}$;
        **end**
        refresh the index of R, U and L starting with 1;
        reset $min_u = u_1$ and $max_l = l_1$;
    **else**
        set $min_u = min(min_u, u_i)$ and $max_l = max(max_l, l_i)$;
    **end**
**end**

Algorithm 1 - The algorithm for decomposition path polygon P.

Therefore, we conclude that the guard γ must be placed within the orthogonal shadow $os_t$ in order to cover the tooth edge t. We demonstrate in the algorithm that the number of guards determined is sufficient to cover the entire polygon, eliminating the need for any additional guards. Please refer to Fig. 5, where certain orthogonal shadows of tooth edges from the upper chain intersect with orthogonal shadows of tooth edges from the lower chain. In such cases, we place a guard at the intersection to reduce the overall number of guards required. The orthogonal shadow of each tooth edge intersects with the corridor.

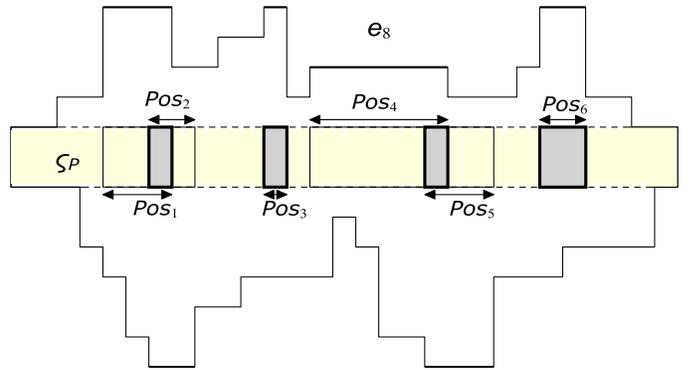

Fig. 5 - balanced polygon and its components

$$Pos_i = os_i \cap \varsigma \qquad (2)$$

Let Pos be the series of $Pos_i$. It represents the positions where guards must be placed. It is necessary to have one guard in each of them. We can identify the positions that minimize the number of guards required within a balanced polygon. The algorithm achieves this in linear time, corresponding to the number of vertices. In Algorithm 2, Pos represents the optimal guard positions, while the variable GN indicates the cardinality or the number of guards in the optimal guard set.

**Lemma 4:** The minimum number of robots required to cover a balanced polygon P is equal to GN, which is obtained through Algorithm 2.

**Proof.** Let's assume that GN guards are sufficient to guard the entire polygon P. Using Lemma 3, we can prove that this number of guards is necessary to guard the tooth edges of P. Each area $Pos_i$ in Pos is a subset of an r-star sub-polygon, meaning that if we decompose P into r-star sub-polygons, the kernels of every r-star sub-polygon contain at least one point from the elements of Pos. Therefore, the entire P is covered by these GN guards and their respective positions. ∎

*C. Efficiency and Time Complexity*

Let's analyze the efficiency and time complexity of our solution. We will explain why our algorithm operates in O(n) time, where n represents the size of the input (path polygon P with n vertices). To begin with, we need to decompose the polygon P into balanced parts using Algorithm 1. The vertical decomposition and the process of finding optimal balanced orthogonal parts can be solved in linear time, denoted as O(n). This is because the number of rectangles formed during the decomposition is on the order of O(n). Once the polygon is decomposed, the problem is divided into subproblems. Each subproblem involves finding the minimum guard set for the obtained balanced sub-polygons. For this purpose, we utilize Algorithm 2, which is designed specifically for guarding monotone parts. The time complexity of solving each subproblem is linear, relative to its size. The total time required to solve all the subproblems is O(n). This is because the sum of the vertices in all the obtained balanced parts is proportional to the size of the input, n. Consequently, Algorithm 2 can be executed in O(n) time for all the balanced parts.

As a result, our entire solution operates in O(n) time, efficiently handling all computations. Finally, GN represents the optimal number of guards needed to cover the path polygon P. Hence, we have demonstrated that our geometric algorithm can find the minimum number of guards required for an orthogonal path polygon P with n vertices, using r-guards, in O(n) time.

---

**Data:** $E_L$ and $E_U$ of the polygon

**Result:** GN and Pos

Set GN = 0 and Pos *is* empty;

Set $e_{min}$ = the lowest horizontal edge of $E_U$;

Set $e_{max}$ = the highest horizontal edge of $E_L$;

**foreach** edge $e_i$ belongs to $E_L$ **do**

    **if** Interior angles of right($e_i$) and left($e_i$) are equal to $\frac{\pi}{2}$ **then**

        $A_i$ = (x(left($e_i$)), y($e_{max}$));

        $B_i$ = (x(right($e_i$)), y($e_{min}$));

        Set $Pos_i$ = rectangle spanned by $A_i$ and $B_i$;

        Set $Pos_L$ = $Pos_L \cup Pos_i$ ;

        GN = GN + 1;

    **end**

**end**

**foreach** edge $e_i$ belongs to $E_U$ **do**

    **if** Interior angles of right($e_i$) and left($e_i$) are equal to $\frac{\pi}{2}$ **then**

        $A_i$ = (x(left($e_i$)), y($e_{max}$));

        $B_i$ = (x(right($e_i$)), y($e_{min}$));

        Set $Pos_i$ = rectangle spanned by $A_i$ and $B_i$;

        Set $Pos_U$ = $Pos_U \cup Pos_i$ ;

        GN = GN + 1;

    **end**

**end**

Merge $Pos_L$ and $Pos_U$ as Pos;

**foreach** $pos_i$ belongs to Pos **do**

    **if** $Pos_i \cap Pos_{i+1}$ is not empty **then**

        $Pos_i$ = $Pos_i \cap Pos_{i+1}$;

        Pos = Pos - $Pos_{i+1}$;

        GN = GN - 1;

    **end**

**end**

Algorithm 2 - The algorithm for optimal guarding in a balanced polygon

---

## IV. CONCLUSION

In this study, our focus was on addressing the problem of finding the minimum number of r-guards required to secure an orthogonal pathway. The goal is to ensure that all points in the pathway are orthogonally visible from at least one guard in G. We have presented an exact and optimal algorithm for solving the guard set problem specifically for pathways. Our algorithm achieves this in linear time, with a time complexity proportional to the number of sides of the pathway, denoted as n. Furthermore, the space complexity of our algorithm is also $O(n)$. The significant outcome of our research is the improvement in the time complexity of the orthogonal art gallery problem for pathways, reducing it from $O(n^{17} poly \log n)$ −time [18] to linear time complexity. As for future work, we aim to further explore this problem and extend our algorithm to handle various types of simple polygons, both with and without holes. Our presented algorithm demonstrates favorable time and space complexity, both being of the order O(n), which currently stands as the best complexity achievable for pathways.

In conclusion, our study provides an efficient and optimal algorithm for finding the minimum number of r-guards in orthogonal pathways. The application of our approach significantly improves the time complexity, offering promising avenues for future research and practical implementations in securing various types of pathways.